\documentclass[conference,a4paper]{IEEEtran}
\IEEEoverridecommandlockouts
\usepackage[labelformat=simple]{subfig}

\usepackage[defaultbib]{bibtopic}
\usepackage{cite}

\usepackage[labelsep=colon]{caption}
\usepackage{amsmath,amssymb,amsfonts}
\usepackage{algorithm}
\usepackage{algpseudocode}
\usepackage{graphicx}
\usepackage{textcomp}
\usepackage{xcolor}
\usepackage{color, colortbl}
\usepackage[defaultbib]{bibtopic}
\def\BibTeX{{\rm B\kern-.05em{\sc i\kern-.025em b}\kern-.08em
    T\kern-.1667em\lower.7ex\hbox{E}\kern-.125emX}}
\begin{document}

\title{Let's Predict Who Will Move to a New Job}

%\title{Let's Predict Who Will Move to a New Job}
%Let's Predict Who Will Move to a New Job

\author{\IEEEauthorblockN{Rania Mkhinini Gahar}
\IEEEauthorblockA{\textit{OASIS Laboratory} \\
\textit{ENIT}\\
\textit{University of Tunis El Manar}\\
Tunis, Tunisia \\
rania.mkhininigahar@enit.rnu.tn}
\and
\IEEEauthorblockN{Adel Hidri}
\IEEEauthorblockA{\textit{Computer Department, Deanship } \\
\textit{of Preparatory Year and Supporting Studies,}\\
\textit{Imam Abdulrahman Bin Faisal University}\\
Dammam, Saudi Arabia \\
abhidri@iau.edu.sa}
\and
\IEEEauthorblockN{Minyar Sassi Hidri}
\IEEEauthorblockA{\textit{Computer Department, Deanship} \\
\textit{of Preparatory Year and Supporting Studies,}\\
\textit{Imam Abdulrahman Bin Faisal University}\\
Dammam, Saudi Arabia \\
mmsassi@iau.edu.sa}
}
\maketitle

\begin{abstract}
Any company's human resources department faces the challenge of predicting whether an applicant will search for a new job or stay with the company. In this paper, we discuss how machine learning (ML) is used to predict who will move to a new job. First, the data is pre-processed into a suitable format for ML models. To deal with categorical features, data encoding is applied and several MLA (ML Algorithms) are performed including Random Forest (RF), Logistic Regression (LR), Decision Tree (DT), and eXtreme Gradient Boosting (XGBoost). To improve the performance of ML models, the synthetic minority oversampling technique (SMOTE) is used to retain them. Models  are assessed using decision support metrics such as precision, recall, F1-Score, and accuracy.
\end{abstract}


\section{Introduction}
A number of factors, including job market competition and personal preferences, lead to people changing jobs over the course of their careers.

However, changing jobs is a difficult decision that may be influenced by a variety of elements, including pay, job description, and location. A successful professional career requires making smooth job changes.

The objective of this work is to accurately predict if an applicant will move to a new job or not using supervised Machine learning (ML) models \cite{AlsaifHFEH22,AlsaifHEFA22,AmmarHH20}.

To evaluate the model, several different implementations of classification were compared to determine which model suits
this type of data best. These were trained on a subset of data originating from available
person profiles collected through web scraping.

The different steps of our approach are as follows:
\begin{itemize}
    \item Data preprocessing (Data cleaning): in this step, null values are removed from the training dataset. Input values have been changed to  specific required data types. Categorical variables are encoded to dummy variables.
    \item Model Building: in this step, several MLAs have been performed including Random Forest (RF), Logistic Regression (LR), Decision Tree (DT), and eXtreme Gradient Boosting (XGBoost). Finally, SMOTE is used to improve the performance of MLAs.
    \item Model evaluating: in this step, models are assessed using decision support metrics such as precision, recall, F1-Score, and accuracy.
    \end{itemize}

The remainder of this paper is organized as follows. The methodology adopted is detailed in section II. Section III evaluates the proposed predictive model. Conclusions and directions for future work are given in Section IV.

 \section{Methodology}
MLAs expect inputs or features to be in a single numeric vector. Similarly, the value to be predicted (label), especially when dealing with categorical data, must be encoded \cite{AmmarHH20}. Thus, one of the objectives of data preparation is to obtain the data in the format expected by the MLAs

\subsection{Human Resources data}
Publicly available Human Resources (HR) data have been used in this work.  The data has 10 features. Data features include the \textit{City Development Index}, the \textit{Gender}, the \textit{Relevant Experience}, the \textit{Enrolled University}, the \textit{Education Level}, the \textit{Major Discipline}, the \textit{Total Years of Experience}, the \textit{Company Size}, and the \textit{Company Size}. The last feature is the \textit{Target} which indicates if the employee seeks a new job or not (0 or 1).  The employees are divided into two classes. The first class contains all those employees who want to move to a new job and the second class consists of all those who did not seek a new job.

Table \ref{test} presents the number of instances used to train and test the model.

            \begin{table}[ht!]
            \centering
                       \caption{Training and Test data}\label{test}
            \vspace{0.3cm}
            \begin{tabular}{cc}
            \hline
            \bf \#Training Data (80\%)&  \bf \#Test Data (20\%)\\
            \hline\hline
          7164 & 1791  \\
            \hline
            \end{tabular}
            \end{table}

Information about the classes (Job seekers) and Non-job seekers) from test Data presented in Table \ref{test} is summarized in Table \ref{train}.
            \begin{table}[ht!]
            \centering
                        \caption{Data description}\label{train}
            \vspace{0.3cm}
            \begin{tabular}{cc}
            \hline
            \bf \#Non job seekers&  \bf \#Job seekers\\
            \hline\hline
          1511 & 280  \\
            \hline
            \end{tabular}
            \end{table}

\subsection{Data encoding}
A categorical variable takes on values called categories, modalities, or levels that have no quantitative meaning. For example, the gender of an individual is a categorical variable with two (or more) modalities: male and female.  Most categorical variables are nominal. These variables are used to categorize and label attributes. Variables contain different values, and each value represents a separate category.

Many MLAs are unable to deal with categorical variables. It is therefore important to encode the data in an appropriate form in order to be able to preprocess these variables. As you need to fit and evaluate your model, you need to code the categorical data and convert all input and output variables to numeric values. Thus, the model will be able to understand and extract the information that generates the desired result. A different set of data varies depending on the number of possible values.

Moreover, the way in which this transformation is carried out is very important. Indeed, the coding of categorical variables generally harms the performance of learning algorithms. One code may be better than another. For example, RF model struggles to capture information from categorical variables with a large number of categories if they are processed with the one-hot encoding technique.

This is how more specific learning algorithms such as XGBoost came into existence.

We have used different methods and tricks to manage the categorical variables present in the dataset used, namely:
\begin{itemize}
    \item \textbf{One-hot encoding}: It consists of coding each categorical variable with different Boolean variables (also called dummy variables) which take the values 0 or 1, indicating whether a category is present in an observation.

Consider a categorical variable $X$ which admits $K$ modalities $m_1$, $m_2$,..., $m_K$. One hot encoding consists of creating $K$ indicator variables, i.e. a vector of size $K$ which has 0s everywhere and a 1 at position $i$ corresponding to modality $m_i$. The categorical variable, therefore, is replaced with $K$ numerical variables.

Some algorithms, in particular some implementations of decision tree (DT) forests, fail to make the best use of the information contained in these variables when this number of modalities is too large.
\item \textbf{Reduction in the number of modalities}: Business knowledge can help reduce the number of modalities. Indeed, an understanding of the categories can allow them to be grouped effectively. A natural grouping is done when the modalities are hierarchical, that is to say, it is possible to define a new category that includes other categories. Suppose a variable whose categories are the districts of a city: these categories can for example be grouped by district, i.e. the districts of the same district will have the same modality. This is a fairly common case. However, these groupings can introduce a bias into the model \cite{3565266}.

A second way to get away with a high number of categories is to try to merge the categories with low counts. Modalities that appear very infrequently in the data can be combined. A frequency table of the modalities is drawn up, and those whose frequency is below a certain threshold are put together in the same \textit{other} category, for example. Then, a one-hot encoding can be applied to the new variable.

\item \textbf{Impact encoding}: When the number of categories becomes very large, encoding by dummy variables can become inconvenient. An alternative method to clustering or truncation of categories consists in characterizing the categories by the link they maintain with the target variable $y$: this is the encoding impact \cite{GNAT20213542}.

This method is also known under the names: likelihood encoding, target coding, conditional probability encoding, and weight of evidence.

For a regression problem with target variable $y$, let $X$ be a categorical variable with $K$ categories  $m_1$, $m_2$,..., $m_K$. Each $m_K$ category is encoded by its impact value:
\begin{equation}
    impact(m_k)=E[y|X=m_k]-E[y]
\end{equation}

$E[y|X=m_k]-E[y]$ corresponds to the expectation of the target $y$ knowing that the variable $X$ is fixed to the modality $m_k$. For a training set of size $n$ containing samples $\{(x_i, y_i)\}$ independent and identically distributed, the estimator of this expectation is the mean of the values of $y_i$ for which the modality $x_i$ is equal to $m_k$:

\begin{equation}
    E[y|X=m_k]=\frac{1}{n_k}\sum_{i \in S_k}y_i
\end{equation}

where $S_k$ is the set of indices $i$ of the observations such that $x_i$ is equal to $m_k$ and $n_k$ the cardinality of this set.

The estimator of the expectation of $y$ is simply its empirical mean:

\begin{equation}
    E[y]=\frac{1}{n}\sum_{i}^ny_i
\end{equation}
\item \textbf{Embedding methods}: This method uses deep learning techniques; it draws its inspiration from models like word2vec on textual data and which gives very impressive results \cite{goldberg2014word2vec}.

This involves creating a representation of each modality of a categorical variable in a numeric vector of fixed size. The use of embeddings allows among other things a reduction of the dimensionality since the size of the vector e can be chosen very small compared to the number of modalities.

Concretely, obtaining these embeddings is done by training a neural network (often a multilayer perceptron) with only the categorical variables as input. First, a one-hot encoding is applied to the variable in order to be input to the network. Generally, one or two concealed coats are sufficient. The first hidden layer has e neurons. The network is then trained on the same task as that initially defined. Then, the output of the first hidden layer then constitutes the vector of embeddings. This vector is then concatenated to the initial data. This data is then used in the fitting of the final model. There are various variations in the literature on how to obtain these embeddings. In addition, nothing prevents putting more than two hidden layers and retaining the output of the second rather than that of the first. Furthermore, the network can be trained on a task other than the initial task.
\end{itemize}

\subsection{Handling Imbalanced Data with SMOTE}
Imbalanced datasets hinder the predictive capability of ML models
due to the biased towards the majority class \cite{j.ins.2018}.

The preprocessed dataset is imbalanced with a bias towards the majority class. To optimize the predictive capability of the ML
models and enhance the generalizability of this study, the preprocessed
(imbalanced sample) data was undersampled and oversampled to create a balanced
distribution.

Undersampling involves removing cases of the majority class and oversampling involves adding instances to the minority class \cite{Kotsiantis}. Undersampling and oversampling have associated benefits and costs.
Undersampling can increase accuracy by decreasing the complexity of the dataset
\cite{Faraoun2007}. Conversely,
undersampling can hinder predictive performance by using a diminutive dataset,
compared to the full sample \cite{j.ins.2018}. Oversampling enhances
predictability through increased data but can result in over-fitting if observations are
duplicated \cite{Weiss2007}.

It is common in machine learning when one of the classes has a much greater or lesser number of observations than the other classes. In this case, the distribution of data is unbalanced. Taking class distribution into account is not possible with MLAs since they increase accuracy by reducing error. Detecting fraud, anomalies, and facial recognition are common examples of this problem.

The majority class bias in standard ML techniques like the DT and LR tends to exclude minorities. Therefore, the minority class is often misclassified as a majority class, due to their tendency to predict only the majority. We are more likely to see negligible or very low recall for the minority class if we have an unbalanced distribution of data in our dataset.

There are mainly 2 widely used algorithms to handle unbalanced class distribution: SMOTE (Synthetic Minority Oversampling Technique) and Near Miss Algorithm.

Oversampling SMOTE aims to balance the distribution of classes by randomly increasing examples of minority classes by reproducing them \cite{1622407}.

The algorithm \ref{alg:cap} illustrates the diffrent steps of SMOTE.

\begin{algorithm}
\caption{SMOTE Algorithm}\label{alg:cap}
\begin{algorithmic}[1]
\State Definition of the set of minority classes $A$
\For{$x \in A$}
\State $k$ nearest neighbors of $x$ -> Euclidean distance between $x$ and all the other samples of the set $A$.
\EndFor
\State Adjust the sampling rate $N$ according to the unbalanced proportion.
\For{$x \in A$}
\State Randomly choose $N$ examples (i.e. $x_1$, $x_2$,...,$x_n$) among its $k$ nearest neighbors
\State Build the set $A_1$.
\EndFor
\For{$x_k \in A_1$ $(k=1$, $2$, $3$,...,$N)$}
\State Generate a new example:
$x^{'} = x + rand(0, 1) * \mid x - x_k \mid$
where $rand(0, 1)$ represents the random number between 0 and 1.
\EndFor
\end{algorithmic}
\end{algorithm}

The main idea of the NearMiss methods is to keep a set of points of the majority class which are close to the points of the minority class, in order to better represent the border which separates $C0$ and $C1$ \cite{TANIMOTO2022117130}.

The NearMiss-1 method selects the elements of the majority class which have the smallest average distance with respect to the $k$ closest examples of the minority class ($k$ being a tuning parameter to be chosen by the user). The NearMiss-2 method retains the cases of the majority category which have the smallest average distance with respect to the $k$ points farthest from the minority category. Finally, the NearMiss-3 method keeps the $k$ examples of class $C0$ which are the nearest neighbors of each element of class $C1$ \cite{TANIMOTO2022117130}.

The basic intuition about how quasi-neighborhood methods work is presented in algorithm \ref{alg:cap2}.

\begin{algorithm}
\caption{NearMiss Algorithm}\label{alg:cap2}
\begin{algorithmic}[1]
\State Calculate the distances between the majority class and the instances of the minority class.
\State Selects the next $n$ instances of the majority class that have the smallest distances with those of the minority class.
\If{There are $k$ instances in the minority class}
\State The closest method will result in $k*n$ instances of the majority class.
\EndIf
\end{algorithmic}
\end{algorithm}

\subsection{Model Building}
\begin{figure}[ht!]
\centerline{\includegraphics[width=9cm]{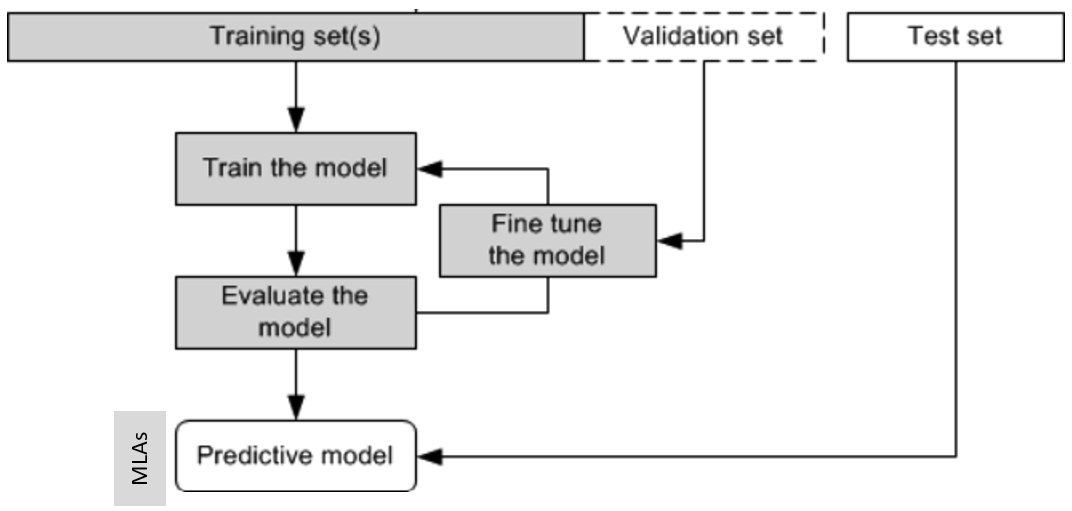}} %angle=90,width=9.5cm,height=21cm
\caption{Predictive modelling process.}\label{fig44hhh}
\end{figure}

Fig. \ref{fig44hhh} shows the predictive modeling process. Let's look at these MLAs used to develop the predictive model.

\begin{itemize}
\item LR \cite{lr}:  LR is a statistical model used to study the relationships between a set of qualitative variables $X_i$ and a qualitative variable $Y$. It is a generalized linear model using a logistic function as a link function.

\item DT \cite{dtdt}: DT are among the most widely used non-parametric supervised learning methods in classification and regression. On the one hand because of their algorithmic simplicity and on the other hand, because of the ease of interpreting them and explaining the results generated. Decision trees are built through an algorithmic approach and can be viewed as a tree with rules that identify ways to split a data set. The goal is to create a model that predicts the value of a target variable by learning the decision rules.
\item RF \cite{breiman2001random}:  This classification algorithm reduces the variance of predictions from a single decision tree, thereby improving their performance. For this, it combines many decision trees in a bagging-type approach.
\item XGBoost \cite{2939785}:
XGBoost is an optimized distributed gradient boosting method. In spite of the fact that Gradient Boost methods are sequential algorithms, XGBoost uses multithread processing to search in parallel for the best split between the features. When compared to other Gradient Boost method implementations, XGBoost performs well due to the utilization of multithreading.
\end{itemize}

\section{Model Evaluation}
As for any ML algorithm, we need to be able to evaluate the performance of the used MLAs in order to decide which algorithm fits our situation best. The following metrics are used in our approach \cite{AlsaifHH22,FerjaniHF22}:

\begin{itemize}
\item Confusion matrix.
\item Precision.
\item Recall.
\item F1 - score.
\item Accuracy.
\end{itemize}

The goal is to predict employees who want to change jobs, so we have two classes (0 for non-job seekers, and 1 for job seekers).

\begin{figure}[ht!]
\centerline{\includegraphics[width=8.7cm]{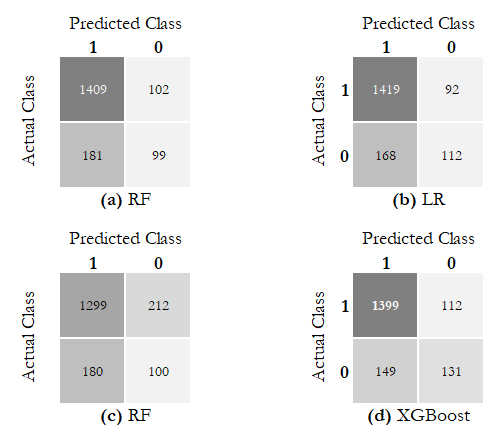}} %angle=90,width=9.5cm,height=21cm
\caption{Confusion Matrix without data balancing.}\label{fig44}
\end{figure}

Fig. \ref{fig44} shows the confusion matrix for RF (Fig. \ref{fig44}(a)), LR (Fig. \ref{fig44}(b)), DT (Fig. \ref{fig44}(c)), and XGBoost (Fig. \ref{fig44}(d)) without data balancing (SMOTE).

\begin{figure}[ht!]
\centerline{\includegraphics[width=8.7cm]{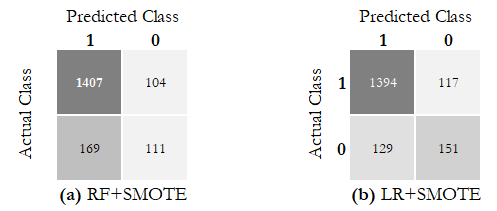}} %angle=90,width=9.5cm,height=21cm
\caption{Confusion Matrix with data balancing.}\label{fig45}
\end{figure}

Fig. \ref{fig45} shows the confusion matrix for LR (Fig. \ref{fig45}(a)) and RF (Fig. \ref{fig45}(b)) with data balancing (SMOTE).

According to Fig. \ref{fig45}, SMOTE significantly improves the prediction performance. SMOTE-LR  outperforms the competition on all metrics, particularly recall (56.34\%) and accuracy (86.26\%).

Table \ref{tabresult} presents the model performance given recall, precision, F1-score, and accuracy values.

  \begin{table}[ht!]
\begin{center}
\caption{Model performance.}
\label{tabresult}
\scriptsize{
\begin{tabular}{lcccc}
\hline
\textbf{MLA}     & \textbf{Precision} & \textbf{Recall} & \textbf{F1-score} &\textbf{Accuracy} \\
\hline\hline
RF& 35.4\%      &50.24\%    &41.53\%&84\%\\
LR& 40\%        &54.9\%     &46.28\%&85.48\%\\
DT& 35.71\%     &32.05\%         &33.78\%&78.11\%\\
XGBoost&46.8\%  &53.9\%     &50.09\%&85.42\%\\
SMOTE-RF&35.71\% &51.62\%    &42.22\%&84.75\%\\
SMOTE-LR&53.9\% &\cellcolor{lightgray}56.34\% &55.09\%&\cellcolor{lightgray}86.26\%\\
\hline
\end{tabular}
}
\end{center}
\end{table}

According to Table \ref{tabresult}, SMOTE-LR gives the best recall and Accuracy.

\section{Conclusion}
In this paper, we tried to predict from HR data who will move to a new job. We performed 6 ML models. The accuracy of each algorithm is evaluated.

The classification rate is assessed using decision ML metrics including recall, precision, F-1 score, and accuracy. As a general rule, most models perform well if the proportions of the classes in a dataset are relatively similar.

Since MLAs struggle to correctly identify the minority class, slight class imbalances were well managed using SMOTE.  SMOTE-LR can be extremely useful for HR department as it has the highest Recall.

In the future direction, we will apply the Convolutional Neural Network-based deep learning (CNN-based DL) model to predict who has the attrition to move to a new job or not.

\section*{References}
\bibliographystyle{IEEEtran}
\bibliography{myBibFile}
\begin{btSect}{myBibFile}		
\btPrintCited			
\end{btSect}
\end{document}